\documentclass[conference,a4paper]{IEEEtran}
\IEEEoverridecommandlockouts
\usepackage{cite}
\usepackage{amsmath,amssymb,amsfonts}
\usepackage{algorithmic}
\usepackage{graphicx}
\usepackage{textcomp}
\def\BibTeX{{\rm B\kern-.05em{\sc i\kern-.025em b}\kern-.08em
    T\kern-.1667em\lower.7ex\hbox{E}\kern-.125emX}}

\usepackage{booktabs}
\usepackage{multirow}
\usepackage{bm}
\usepackage{float} %figure inside minipage
\usepackage{pifont}
\usepackage{enumitem}
\usepackage{url}
\usepackage{sidecap}

\newcommand{\cmark}{\ding{51}}%
\newcommand{\xmark}{\ding{55}}%
\usepackage[table]{xcolor}
\definecolor{gray1}{rgb}{0.84,0.84,0.84}
\definecolor{gray2}{rgb}{1,0.89,0.75}
\definecolor{clr3}{rgb}{0.95,0.95,0.95}
\definecolor{clr4}{rgb}{0.96,0.96,0.86}

\usepackage{xspace}

\makeatletter
\DeclareRobustCommand\onedot{\futurelet\@let@token\@onedot}
\def\@onedot{\ifx\@let@token.\else.\null\fi\xspace}

\def\eg{\emph{e.g}\onedot} 
\def\ie{\emph{i.e}\onedot}

\makeatother

\begin{document}

\title{Question-Agnostic Attention for Visual Question Answering
}

\author{\IEEEauthorblockN{Moshiur R Farazi}
\IEEEauthorblockA{\textit{Australian National University} \\
\textit{and Data61, CSIRO}\\
Canberra, Australia \\
moshiur.farazi@anu.edu.au}\\
\and
\IEEEauthorblockN{Salman Khan}
\IEEEauthorblockA{\textit{Inception Institute of AI} \\
\textit{and Australian National University}\\
Abu Dhabi, UAE \\
salman.khan@anu.edu.au}\\
\and
\IEEEauthorblockN{Nick Barnes}
\IEEEauthorblockA{\textit{Australian National University} \\
Canberra, Australia \\
nick.barnes@anu.edu,au}
% \and
% \and
% \and
}

\maketitle

\begin{abstract}
Visual Question Answering (VQA) models employ attention mechanisms to discover image locations that are most relevant for answering a specific question. 
For this purpose, several multimodal fusion strategies have been proposed, ranging from relatively simple operations (\eg linear sum) to more complex ones (\eg Block \cite{Ben_2019_AAAI}). 
The resulting multimodal representations define an intermediate feature space for capturing the interplay between visual and semantic features, that is helpful in selectively focusing on image content. 
In this paper, we propose a question-agnostic attention mechanism that is complementary to the existing question-dependent attention mechanisms.
Our proposed model parses object instances to obtain an `object map' and applies this map on the visual features to generate Question-Agnostic Attention (QAA) features.
In contrast to question-dependent attention approaches that are learned end-to-end, the proposed QAA does not involve question-specific training, and can be easily included in almost any existing VQA model as a generic light-weight pre-processing step, thereby adding minimal computation overhead for training.
Further, when used in complement with the question-dependent attention, the QAA allows the model to focus on the regions containing objects that might have been overlooked by the learned attention representation.
Through extensive evaluation on VQAv1, VQAv2 and TDIUC datasets, we show that incorporating complementary QAA allows state-of-the-art VQA models to perform better, and provides significant boost to simplistic VQA models, enabling them to performance on par with highly sophisticated fusion strategies.
\end{abstract}


\section{Introduction}
\label{sec:intro}
An attention mechanism in a VQA system identifies the relevant visual  information to intelligently answer a given question. Therefore, attention is central to recent state-of-the-art VQA models. Existing VQA models generally use grid-level or object-level convolutional features to \textit{learn} an attention distribution over the given image. In the former case, this attention is dispersed over the spatial grid \cite{antol2015vqa, fukui2016multimodal, yang2016stacked} while in the later case, attention is applied on a set of object proposals \cite{Anderson2017up-down, Ben_2019_AAAI}. Recent best performing methods combine the strengths of both these approaches to obtain attention maps with a better context \cite{Farazi_2018_BMVC,lu2018co}.

%%%%%%%%%%%%%%%%%%%%%%%%%%%%%%%
\begin{figure}
\centering
% \fbox{\rule{0pt}{2in} \rule{.9\linewidth}{0pt}}
\includegraphics[width=.75\linewidth]{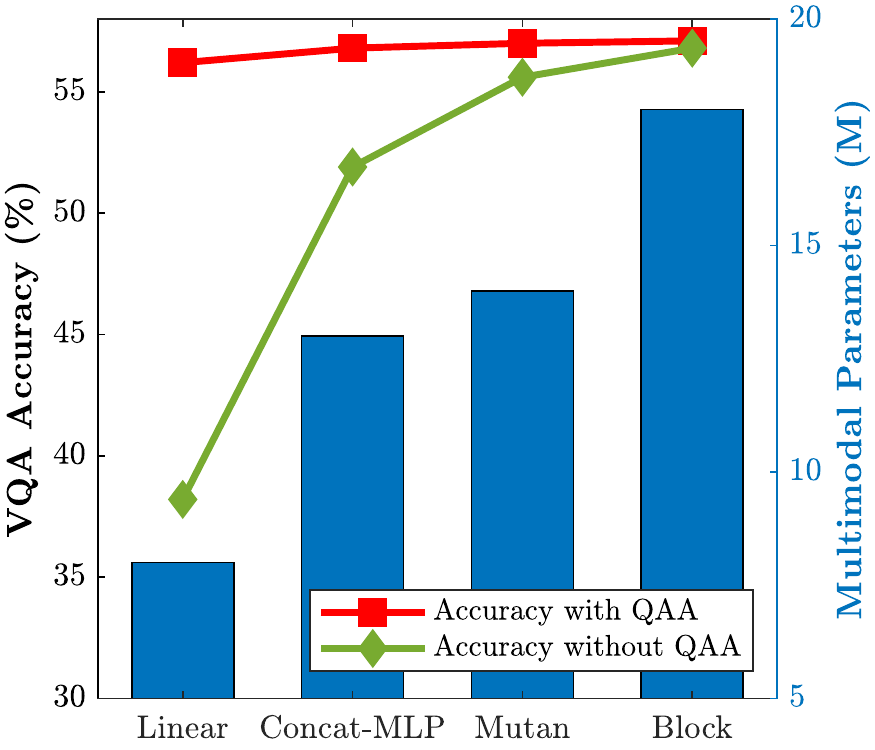}
% \vspace{-0.85em}
  \caption{{A comparison of various multimodal fusion schemes for VQA evaluated on VQAv2 validation dataset. In general, methods with low-parametric complexity (such as linear sum, concatenation followed by MLP) deliver low performance compared to more sophisticated ones (\eg Mutan \cite{ben2017mutan}, Block \cite{Ben_2019_AAAI}). Using our proposed Question-Agnostic Attention, we observe a consistent boost for all fusion mechanisms. The improvement is especially more pronounced for simple models, bringing them on par with highly sophisticated methods.}}
\label{fig:teaser}
\end{figure}
%%%%%%%%%%%%%%%%%%%%%%%%%%%%%%%%

Learned attention mechanisms have been shown to significantly enhance the performance of VQA systems. However, learning attention on dense grid- and object-level features is a computationally demanding task that results in increased model complexity. Furthermore, learned attention is tuned for a specific dataset and thereby fails to generalize well to novel scenarios. To address these problems, we undertake a tangential path and propose a Question-Agnostic Attention (QAA) approach that is independent of a given question. Our approach is based on the insight that questions generally relate to the state, number, type and actions of the `\textit{objects}' present in an image and their `\textit{mutual relationships}'. Therefore, we propose to use an object parsing module to generate question-agnostic attention maps based only on the given images. This attention generation procedure acts as a simple pre-processing step that encodes salient instance-centric visual cues (\eg location, shape) and object-relationship information which in turn leads to a performance boost for all evaluated models  and difficult question types (\eg \textit{`What sports are they playing?'}, \textit{`What kind of furniture is in the picture?'}).

Furthermore, several efforts in VQA literature show the importance of object-aware visual attention for improved VQA \cite{lu2016hierarchical, Anderson2017up-down, Farazi_2018_BMVC, Farazi_IMAVIS_2020} which emphasizes the notion that better localization of object instances results in higher VQA accuracy. However, these attention procedures are learned on top of object proposals while we propose an attention approach with minimal training cost. Our approach uses instance segmentation to generate an \textit{object map} on the spatial image grid in a bottom-up fashion that is demonstrated to improve performance for simple as well as complex VQA models.

Our results provide an interesting  perspective on VQA showing that question-agnostic attention can help achieve competitive VQA performance and provides complementary information for existing VQA models, that results in notable performance gain. In an extreme case, when we apply a fixed attention map computed from a prior based on the training data, the VQA model still performs on par with existing models with learned attention. \textit{Firstly,} this highlights the performance-complexity trade-off that is offered by recent multimodal fusion mechanisms for VQA task.

Our results show that even with very simple multimodal operations, a VQA model can perform as well as more sophisticated models if question-agnostic attention is used. \textit{Secondly,} the performance improvement across all the models illustrates the complementary nature of QAA, that highlights the room for improvement in learned `question-aware' attention. \textit{Finally,} the relatively stronger improvement for simpler models shows that the information learned with QAA features is somewhat similar to the high-order representation modelling through complex multimodal fusion techniques. 

The main contributions of our paper include:
% \begin{itemize}\setlength{\itemsep}{-.25em}
\begin{itemize} %[topsep=-5pt,noitemsep]
\item An inexpensive VQA pre-possessing step, dubbed Question-Agnostic Attention (QAA), that localizes object instances in an image irrespective of the question.
\item A modular co-attention architecture that allows any off-the-shelf VQA model to incorporate complementary QAA features.
\item An extensive set of experiments on large scale VQA datasets and the {TDIUC dataset} to showcase the effectiveness of using complementary QAA features,  {e}specially helping simplistic VQA models achieve near state-of-the art performance.
\end{itemize}

%-------------------------------------------------------------------------
\section{Related Works}
\label{sec:related_works}
\textbf{Visual feature extraction:}  State-of-the-art VQA models either use deep CNN based feature extraction networks (\eg ResNet~\cite{he2016deep}) to generate visual features for each grid location in an image \cite{fukui2016multimodal, ben2017mutan, yang2016stacked}, or employ object detectors (\eg. Faster-RCNN~\cite{ren2015faster}) to detect the best object proposals for a given image and extract a corresponding set of object-level features \cite{Anderson2017up-down, Ben_2019_AAAI, zhang2018vqacount}. Attention mechanisms are then applied to selectively consider the relevant information on the spatial grid or the object proposals. Some recent approaches also combine spatial grid and object-level attention to leverage the best of both approaches \cite{Farazi_2018_BMVC, lu2018co}. However, the feature maps generated from the object proposals are discrete and do not encode the spatial relationship between the objects present in the image. Thus there exists a semantic gap since the two sets of approaches look at different kinds of features, one from image level and one from object level. Here, we propose to use the object location information on top of spatial-grid features to bridge this semantic gap. \\
 
\textbf{Attention mechanisms:} For an improved VQA capability, attention has been focused on either or both the image \cite{shih2016look} and the natural language questions \cite{lu2016hierarchical}. Stacked attention networks \cite{yang2016stacked} generate an attention map by recursively attending to salient image details. VQA-HAT \cite{das2017human} studies human attention maps for VQA and quantifies how they correlate with automatically learned attention maps. Task-independent saliency prediction methods have also been used as an attention mechanism for VQA. There exists a strong center-bias in human eye-fixation and saliency based methods \cite{das2017human,judd2009learning}. Our study also confirms this behaviour as we show that the global representation of \textit{object  map} is more concentrated towards the image-center. 

%%%%%%%%%%%%%%%%%%%%%%%%%%%%%%%%
\begin{figure*}
\centering
% \fbox{\rule{0pt}{2in} \rule{.9\linewidth}{0pt}}
\includegraphics[width=.95\linewidth]{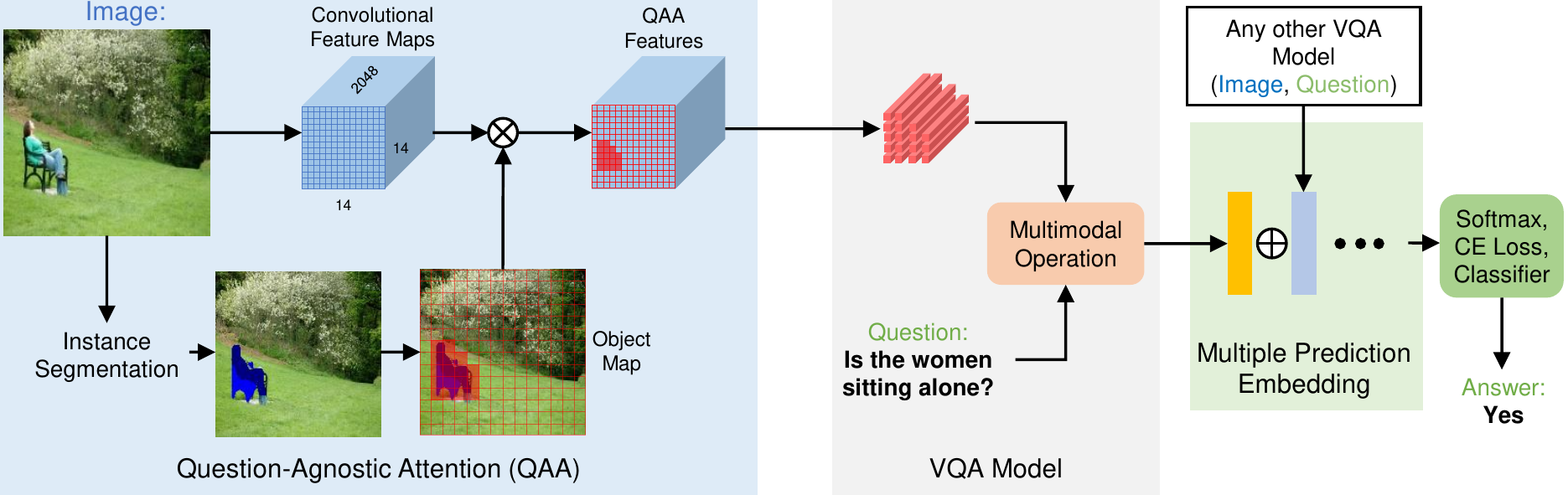}
% \vspace{-0.8em}
  \caption{{Architecture of our Question-Agnostic Attention (QAA) based VQA model. QAA features are generated using instance segmentation (generated by Mask-RCNN) to create a binary \textit{object map} with the same resolution of the convolutional feature map. The \textit{object map} is applied as a mask on the convolutional feature map (generated by ResNet) of the whole image. This `modular attention' with minimal training cost delivers strong improvement while used in complement with existing VQA models on a number of VQA benchmarks.}} 
\label{fig:pipleline}
\end{figure*}
%%%%%%%%%%%%%%%%%%%%%%%%%%%%%%%%

%-------------------------------------------------------------------------
\section{Method}
\label{sec:method}
Given a question $Q$ about an image $I$, an AI agent designed for the VQA task will predict an answer $a^*$ based on the learning acquired from training examples. Benchmark VQA models frame this task as a multi-class classification problem in the candidate answer space, and the models learn to predict the correct answer for a given Image-Question (IQ) pair. This task can be formulated as: 
\begin{align}
    a^* = \underset{\hat{a} \in \mathcal{D}}{\arg\max}\, P(\hat{a}|Q,I; \theta), 
\end{align}
where $\theta$ denotes the model parameters and $a^*$ is predicted from the dictionary of candidate answers $\mathcal{D}$. 

A simplistic VQA model consists of two major parts: (1) Feature extraction module, and (2) Multimodal feature embedding. The \textbf{first} part of the model extracts visual features from an Image $I$ and semantic features from a Question $Q$. The visual features from an image are extracted using deep CNN based object recognition models (\eg ResNet~\cite{he2016deep}) which are pretrained on large-scale image recognition datasets such as ImageNet~\cite{deng2009imagenet}. The image feature map from the last convolution layer of the model is extracted as the visual feature $\bm{v} \in \mathbb{R}^{g \times d_v} $, where $g$ is the index of the spatial location in the image over a coarse grid and $d_v$ is the feature embedding dimension for each spatial grid location. On the other hand, for extracting {the} language feature from {a} Question, each word is fed to a pretrained encoder (\eg GloVe~\cite{pennington2014glove}, Skip-thought~\cite{kiros2015skip}) to get vector embeddings of the question words. These vectors are then passed through a language model which consists of Gated Recurrent Units (GRUs) to generate {a} semantic feature $\bm{q} \in \mathbb{R}^{d_q}$.

In the \textbf{second} part, extracted visual and semantic features are combined into a multimodal representation, which in turn is used to minimize a loss function to predict the correct answer. A VQA model employs a joint embedding function $\Psi(\cdot)$ to merge the extracted features in a common multimodal space. The function $\Psi(\cdot)$ can be a simple fixed function (\eg a linear sum, concatenation followed by MLP) or a complex operation (\eg multimodal pooling \cite{fukui2016multimodal} or fusion \cite{ben2017mutan, Ben_2019_AAAI}). Most importantly, the multimodal embedding is used to selectively attend to visual features using a learned attention mechanism. This attention is derived jointly from the given question and image pair. Different from these attention approaches, we propose a pre-prcoessing step that estimates an attention map \textit{without} considering at-all the input questions. This simple approach with no-training cost surprisingly gives highly compelling results.

Our proposed question-agnostic attention model is illustrated in Fig.~\ref{fig:pipleline}. We first employ an attention mechanism that focuses on different object instances by creating an `\textit{object map}' with which the question-agnostic features are generated (Sec.~\ref{sec:ins_att}). The question-agnostic attention enables the model to focus on arbitrary object shapes and object parts which results in an improved model attention. Instead of {the} original CNN extracted spatial grid feature map, the question-agnostic features are passed through the VQA model where the given language query is used to further refine the visual features. These refined visual features are used to generate final predictions for classification. The modular architecture of our model enables it to combine predictions from other VQA models that aggregates multiple predictions to generate an intelligent answer for the given question (Sec.~\ref{sec:mul_dec_em}).

\subsection{Question-Agnostic Attention}
\label{sec:ins_att}
The input image is passed through a pre-trained instance segmentation module to predict the pixels that correspond to object instances. Notably, we ensure that the pre-trained network has not seen any of the test images for the evaluated datasets and is pre-trained on an altogether different task (\ie instance segmentation as opposed to VQA). These instances have arbitrary shape and size which makes it harder to encode them and computationally infeasible for a VQA model to train with instance-level features. To remedy this, a coarse representation of the object instances is generated by creating a grid of size $g$ over the whole image and the object instances are mapped onto this grid. A binary representation of this grid is called the \textit{object map} $\mathcal{M} \in \mathbb{R}^{g}$, which essentially identifies if an object instance occupies a grid location or not. 

One can learn a non-linear mapping function that maps the object instances to an arbitrary high-dimensional space. However, we adopt a simplistic approach to set the size of the \textit{object map} equal to spatial grid size $g$ for primarily three reasons. \textit{Firstly}, having the grid size equal to the CNN features allows our approach to establish a one-to-one correspondence to the spatial grids of the CNN extracted convolutional feature map, which enables the model to access the visual features of that grid region without requiring another explicit ROI pooling like Faster-RCNN. This avoids expensive computations in our model. \textit{Secondly}, the binary $g$-dimensional \textit{object map} can be applied as a mask to select only the visual features at grid locations that have an object instance with a computationally inexpensive element-wise multiplication between $\bm{v}$ and $\mathcal{M}$ to generate question-agnostic feature $\bm{v}^{\mathcal{M}} \in \mathbb{R}^{g \times d_v}$. \textit{Finally}, as the question-agnostic features have the same size as CNN extracted visual features, it can be easy for any VQA model to incorporate the question-agnostic features by only adjusting the size of \textit{object map} equal to the size of CNN spatial grid. Thus, this simplistic approach fashions question-agnostic attention mechanism as an inexpensive pre-processing step that is easily applicable to any CNN based VQA model.

\subsection{Multiple Prediction Embedding}
\label{sec:mul_dec_em}
The modular architecture of QAA enables it to jointly consider predictions from any other VQA model to generate a final prediction vector. In order to further validate the complementary nature of our proposed QAA model, we include a simple spatial attention mechanism, commonly used in most VQA models \cite{fukui2016multimodal, ben2017mutan, Ben_2019_AAAI} to refine the visual features according to the question. In addition to  {a} fixed \textit{object map} used to generate question-agnostic features, this optional module can be used to refine the question-agnostic features according to the question, providing flexibility to incorporate a spatial attention mechanism on top of QAA. We achieve this by calculating a similarity measure between each question-agnostic feature grid location $\bm{v}^{\mathcal{M}}_i \in \mathbb{R}^{d_v}$ and $\bm{q}$ by projecting them in a common space by a joint embedding function $\Psi(\cdot)$. This, in general, represents the relevance of a spatial grid location for answering that input question. This similarity measure is applied as  {a} semantic weighting function, called Spatial Attention $\alpha \in \mathbb{R}^g$, that takes a weighted sum over the spatial grids of input visual features. It can be expressed as:
\begin{equation}
    \Tilde{\bm{v}}^{\mathcal{M}} = \sum_{i=1}^{g} \alpha_{i} \bm{v}^{\mathcal{M}}_i \quad \textrm{where} \;\;
    \alpha_{i} = \textup{softmax} (\Psi (\bm{q},\bm{v}_i))
\end{equation}
where $\Tilde{\bm{v}}^{\mathcal{M}} \in \mathbb{R}^{d_v}$ represents a combination of question-agnostic attention features that are emphasized by the input question. Finally, it undergoes a second multimodal embedding with the question feature $\bm{q}$ to generate a prediction vector $P$ which has the same dimension as the candidate answer dictionary $\mathcal{D}$. {Predictions from any other VQA model can be concatenated with the prediction of our model. The concatenated predictions are passed through a multiple prediction embedding layer that learns to generate a $\mathcal{D}$ dimensional final prediction vector.}

%-------------------------------------------------------------------------
\section{Experiments and Results}
\label{sec:experiments}
In this section, at first. we describe the experimental setup that includes our instance segmentation pipeline, VQA model architecture, dataset and evaluation metric. Then we discuss the findings from our ablative experiments to study effectiveness of our proposed approach in different settings. Finally, we present the qualitative and quantitative results of our model and do a comparative analysis with other state-of-the-art models. 

\textbf{VQA dataset:} Firstly, we evaluate our QAA model on two large scale benchmark VQA datasets, namely VQAv1 \cite{antol2015vqa} and VQAv2 \cite{Goyal_2017_CVPR}. Among the two datasets, VQAv2 contains complementary question-answer pairs which mitigates language bias present in the VQAv1 dataset, making VQAv2 a more challenging test setting. Both versions of the VQA dataset contain over 200K real images sourced from the MS COCO dataset \cite{lin2014microsoft} and placed into respective train/val/test splits. These images are paired with complex open-ended natural language questions and answers. The ground truth answers for train and val split are publicly available, but the test split is not. To evaluate on the test split (both test-dev and test-std), the prediction needs to be submitted to the VQA test server. We perform ablation experiments on validation sets of VQAv1 and VQAv2 (Tab.~\ref{tab:ablation_VQAval}) and compare with other state-of-the-art methods on VQAv2 test-dev and test-std dataset (Tab.~\ref{tab:VQAv2_testset}). Following the standard evaluation strategy, we calculate the accuracy $\hat{a}$ of the predicted answer $a^*$ as $\hat{a} = \mathrm{min} (\# \; \text{of humans answered } a^*/3, 1)$,
% \begin{equation}
% \hat{a} = min \bigg(\frac{\# \; \text{of humans answered } a^*}{3}, 1\bigg),
% \end{equation}
which means that the answer provided by the model is given 100\% accuracy if at least 3 human annotators who helped create the VQA dataset gave the exact answer.

%%%%%%%%%%%%%%%%%%%%%%%%%%%%%%%%%%%%%%%
\begin{table*}[!htp]
    \caption{Comparison of different multimodal operations when using complementary QAA features on VQA datasets. The models are evaluated on validation sets of VQAv1\cite{antol2015vqa} and VQAv2\cite{Goyal_2017_CVPR} dataset, and we report the overall accuracy (higher the better). Models in rows (1)-(3) do not have any spatial attention mechanism whereas the models in rows (4)-(6) learn spatial attention as described in Sec.\ref{sec:mul_dec_em}}
    \centering
    \scalebox{.9}{%@{\extracolsep{3pt}}
    \begin{tabular}{l c c cccc cccc}
    \toprule
    \rowcolor{clr3} &&                      &\multicolumn{4}{c}{VQAv1 Dataset}      &\multicolumn{4}{c}{VQAv2 Dataset} \\
    \cmidrule(lr){4-7} \cmidrule(lr){8-11}
    \rowcolor{clr3}  && Spatial              &\multicolumn{4}{c}{Multimodal Operation}      &\multicolumn{4}{c}{Multimodal Operation} \\                        
    \rowcolor{clr3} & Visual Feature & Attention    & Linear & C-MLP & Mutan  & Block & Linear & C-MLP & Mutan  & Block \\
                                    \cmidrule(lr){2-2}\cmidrule(lr){3-3} \cmidrule(lr){4-5} \cmidrule(lr){6-7} \cmidrule(lr){8-9} \cmidrule(lr){10-11}
    (1)& Spatial Grid (SG)  & \xmark
                            &39.7   &57.2   &56.3   &58.2
                            &38.2   &51.9   &55.6   &56.8\\
    (2)& QAA                & \xmark
                            &41.4   &40.5   &57.3   &58.4
                            &39.7   &53.2   &56.3   &56.5\\
    \rowcolor{clr4} (3) &Ours(SG+QAA)       & \xmark \xmark
                            &\textbf{57.9}     &\textbf{58.3}   
                            &\textbf{57.5}      &\textbf{58.4}
                            & \textbf{56.2}    & \textbf{56.8}
                            & \textbf{57.0}     & \textbf{57.1}\\
                            \cmidrule(lr){4-7} \cmidrule(lr){8-11}
    &   &                   &18.2 $\uparrow$    &1.1 $\uparrow$ &1.2 $\uparrow$ &0.2 $\uparrow$
                            &18.2 $\uparrow$    &4.9 $\uparrow$ &1.4 $\uparrow$ &0.3 $\uparrow$\\
    \midrule
    (4)& Spatial Grid (SG)  & \cmark 
                            &41.8   &60.4   &58.6   &61.2
                            &41.0   &54.4   &57.9   &60.1\\
    (5)& QAA                & \cmark
                            &41.4   &59.6   &57.9   &60.5
                            &37.3   &57.3   &56.5   &59.3\\
    \rowcolor{clr4} (6)&Ours(SG+QAA)        & \cmark \cmark
                            &\textbf{60.6}     &\textbf{60.7}
                            &\textbf{59.2}      &\textbf{61.6}
                            &\textbf{59.1}     &\textbf{59.5}
                            & \textbf{58.2}     &\textbf{60.5}\\
                            \cmidrule(lr){4-7} \cmidrule(lr){8-11}
    &   &                   &18.8 $\uparrow$    &0.3 $\uparrow$ &0.6 $\uparrow$ &0.3 $\uparrow$
                            &18.1 $\uparrow$    &5.2 $\uparrow$ &0.3 $\uparrow$ &0.4 $\uparrow$\\
    \midrule
     & Multimodal Parameters & &8M & 13M &14M &18M &8M & 13M &14M &18M \\
    \bottomrule
    \end{tabular}
    }
    % \vspace{-0.5em}
    % }
    \label{tab:ablation_VQAval}
\end{table*}
%%%%%%%%%%%%%%%%%%%%%%%%%%%%%%%%%%%%%%%

\textbf{TDIUC dataset:} Task Directed Image Understanding Challenge (TDIUC) dataset \cite{kafle2017analysis} consists of $~1.6M$ questions and $~170K$ images sourced from MS COCO and the Visual Genome Dataset. These Image-Question pairs are split into 12 categories and 4 additional evaluation matrices ($1^{st}$ column of Tab.~\ref{tab:TDIUC_results}) which help evaluate a model's robustness against answer imbalance and its ability to answer questions that require higher reasoning capability. We evaluate and perform ablation on TDUIC testset, and report accuracy for all 12 question types along with overall arithmetic mean-per-type (MPT) and harmonic MPT, and overall normalized arithmetic MPT and harmonic MPT in Tab.~\ref{tab:TDIUC_results}.

\textbf{Instance Segmentation:} We employ a pre-trained Mask-RCNN~\cite{he2017mask} model\footnote{\scriptsize \url{github.com/facebookresearch/maskrcnn-benchmark}} to generate instance masks by running inference on the input image. Specifically, the Mask-RCNN model was trained on COCO \textit{train} and the \textit{val-minus-minival} split with a ResNet-50-FPN backbone. Note that although the `training data' (\ie images) of the VQA datasets have an overlap with the `training set' of COCO, none of the test images have been previously seen by the pre-trained model. Also, we do not use any object-level information in our attention map, rather only a simple binary mask showing the location of detected objects is used in our approach. Therefore, our setting has no extra advantage or external supervision compared to other approaches. 

\textbf{Model Architecture:} We use ResNet~\cite{he2016deep} pretrained on ImageNet~\cite{deng2009imagenet} to extract the visual features of an image with dimensions $196 \times 2048$. Here, $g = 196$ which represents the $14 \times 14$ spatial grid corresponding to image regions and $2048$ is the dimension of visual features for each grid location. The language model generates a $d_q = 2400$ dimensional feature for each question in a fashion similar to \cite{fukui2016multimodal, ben2017mutan, yu2018beyond}. The question words are first preprocessed, tokenized and encoded through a embedding layer that consists of GRUs and uses a pretrained skip-thought encoder. For the models \textit{without} the optional spatial attention mechanism, the input visual feature is averaged across the spatial grid to generate a $2048$-d feature vector from the $2048 {\times} 14 {\times} 14$ dimensional feature map and passed on to be jointly embedded with the question feature. On the other hand, the models \textit{with} spatial attention learn to generate $2048$-d feature vector as discussed in Sec.~\ref{sec:mul_dec_em}. Following the VQA benchmark~\cite{antol2015vqa}, the dictionary of candidate answers $\mathcal{D}$ consists of the top $3000$ frequent answers from the respective versions of VQA dataset. A cross entropy loss is minimized to predict the correct answer from the dictionary $\mathcal{D}$. While performing experiments on the TUDIC dataset, dimension of $\mathcal{D}$ is set to $1480$.

\textbf{Baseline Model:} We setup our VQA baseline model with four variants where the model employs different multimodal operations for combining the question and image features. All other setup and hyperparameters are kept exactly the same for fair comparison. Each variant can have the optional spatial attention module. The first two variant of our baseline model incorporate simpler multimodal operation (\ie liner summation and concatenation followed by MLP). The latter two variants use a more complex multimodal operation, namely Mutan~\cite{ben2017mutan} and Block~\cite{Ben_2019_AAAI}, which achieve the state-of-the-art performance for the VQA task, and have a considerably higher number of trainable parameters for multimodal embedding. Mutan and Block operation are implemented using their publicly available code\footnote{\scriptsize\url{github.com/Cadene/block.bootstrap.pytorch}}. The following are the four variants of our baseline model:

\noindent \textit{Linear:} The question and image features are projected onto a common space using fully connected layers and the projected vectors are summed to obtain a joint feature representation. This joint representation is projected to the prediction space $P \in \mathbb{R}^{3000}$ which is then passed through the answer prediction network to generate the final prediction. This can be expressed as: $P' = \omega_P (\omega_{\bm{q}} \bm{q} + \omega_{\bm{v}} \bm{v})$
% \begin{equation}
%     P' = \omega_P (\omega_{\bm{q}} \bm{q} + \omega_{\bm{v}} \bm{v})
%     \label{eq:lin_sum},
% \end{equation}
where $\omega$ represents the fully connected layer weights used for projection.

\noindent \textit{Concat-MLP:} The question and image features are concatenated and passed through a 3-layer MultiLayer Perceptron (MLP) with ReLU activation and dropout to combine the input features. The resulting vector is projected onto the prediction space for answer classification.

\noindent \textit{Mutan:} The Mutan model learns a multimodal interaction between question and image using rank constrained Tucker tensor decomposition~\cite{ben2017mutan}. In this model, the visual and language features are decomposed into three matrices and a core tensor that is somewhat capable of modelling the fully parameterized interaction in the multimodal space.  

\noindent \textit{Block:} It employs block-term tensor decomposition following a super-diagonal fusion framework\cite{Ben_2019_AAAI}. This is the most computationally expensive model that we experiment with and achieves state-of-the-art performance. The complexity of a multimodal operation is inferred by calculating the number of trainable parameters from attended image features, the question embedding, and the answer prediction.

\textbf{Ablation study:} 
In Sec.~\ref{sec:diff_mul_op} we perform ablation to showcase the effectiveness of using complementary question-agnostic attention on VQA models employing different multimodal operation by evaluating on the VQAv2 Valset~\cite{Goyal_2017_CVPR} and the TDIUC testset \cite{kafle2017analysis}. Furthermore, in Sec.~\ref{sec:inf_global_rep} we show that without an explicit \textit{object map} during inference, our model can utilize image independent QAA features generated from a global representation of training examples.

\subsection{Ablation on Different Multimodal Operations}
\label{sec:diff_mul_op}
\noindent \textit{{Simplistic VQA models get a significant performance boost using complementary QAA features and perform on par with the state-of-the-art.}} 

In row (1) of Tab.~\ref{tab:ablation_VQAval}, we report that our baseline VQA model employing state-of-the-art Block fusion achieves $58.4$ and $57.1$ accuracy, whereas with a linear-sum operation, the same model achieves accuracy of only $39.7$ and $38.2$ on VQAv1 and VQAv2 validation sets, respectively. When the Linear model is trained with complementary QAA features, the accuracy increases to $57.9$ and $56.2$ on the VQAv1 and VQAv2 datasets, respectively; performing very close to the state-of-the-art Block model (row (3)). This pattern also exists when these same models include the optional spatial attention module (comparing rows (4) and (6)). The simpler Linear model benefits from using complementary QAA features as it represents a subset of the spatial locations of the whole image that has object instances and encodes visual cues like count, location and attributes which are most important to predict the correct answer. The Linear model with only 8M trainable parameters and relatively simpler multimodal operation cannot learn to identify these visual cues on its own. Thus the performance boost while using complementary QAA feature is more significant (${\sim}18\uparrow$ vs ${\sim}0.5 \uparrow$) for VQA models employing a simplistic multimodal operation (\ie Linear and Concat-MLP) compared to the models employing a more sophisticated fusion operation (\ie Mutan\cite{ben2017mutan} and Block\cite{Ben_2019_AAAI}). Since more complex multimodal operations learn salient visual cues by modeling the interaction between visual and semantic features through significantly more parameters; VQA models employing such complex operations benefit less from using complementary QAA features.  Overall, it can be seen from Tab.~\ref{tab:ablation_VQAval} that all variants of our VQA baseline employing different multimodal operations, with or without optional spatial attention, benefit from using complementary QAA features.

%%%%%%%%%%%%%%%%%%%%%%%%%%%%%%%%%%%%%%%%
\begin{table}[!t]
\caption{Comparison with state-of-the-art single (not ensemble) VQA models with our proposed QAA model, evaluated on VQAv2 Test-dev and Test-std dataset. The models in (1) are trained with spatial grid features and in (2) with Bottom-Up features \cite{Anderson2017up-down}. With QAA, in both cases, our model outperform contemporary VQA models.}
\centering
  \scalebox{.95}{
\begin{tabular}{c c c cccc}
\toprule
\rowcolor{clr3}  &       & \multicolumn{1}{c}{Test-dev}  & \multicolumn{4}{c}{Test-Standard}\\
\rowcolor{clr3} &Model  & All & All 	& Y/N   & Num.  & Other \\ \cmidrule(lr){2-2}\cmidrule(lr){3-3} \cmidrule(lr){4-7}
\multirow{3}{*}{(1)}  
&MCB\cite{fukui2016multimodal}
                                &-      &62.3   &78.8   &38.3   &53.3\\
&Mutan~\cite{ben2017mutan}      &63.2   &63.5   &80.9   &38.6   &54.0\\
\rowcolor{clr4} & Ours(SG+QAA)  &\textbf{64.7}   &\textbf{65.0}   &\textbf{81.8}   &\textbf{43.6} &\textbf{55.4}\\
\midrule
\multirow{3}{*}{(2)}     
&Up-Down\cite{Anderson2017up-down} 
                                &65.3   &65.7   &82.2   &43.9   &56.3\\
&Block~\cite{Ben_2019_AAAI}
                                &66.4   &66.9   &83.8   &45.7   &57.1\\
\rowcolor{clr4} & Ours(BU+QAA)  &\textbf{66.7}    &\textbf{67.0}   &\textbf{83.8}   &\textbf{45.9}   &\textbf{57.1}\\
\bottomrule
\end{tabular}
}
% \vspace{-0.5em}
\label{tab:VQAv2_testset}
\end{table}
%%%%%%%%%%%%%%%%%%%%%%%%%%%%%%%%%%%%%%%

\noindent \textit{Complementary QAA features help answer rare questions more accurately.} In Tab.~\ref{tab:TDIUC_results}, we evaluate our baseline models with and without complementary QAA features on the TDIUC testset and compare it against other state-of-the-art models using spatial grid features. 

The baseline models reported in this table use the optional spatial attention module. We can see that the accuracy for the difficult question categories (\eg Object Utility, Object Presence) increased when using QAA features, and this improvement is more prominent for models using Linear and Concat-MLP operations. Further, for all variants of the baseline model, both versions of Arithmetic and Harmonic MPT improved, and this improvement is more significant for Harmonic MPT and Harmonic N-MPT. This is particularly important as Harmonic MPTs is a more strict metric as it measures the ability of a model to have high scores across `\textit{all}' question-types and it consequently puts an emphasis on lowest performing categories. In the last row of Tab.~\ref{tab:TDIUC_results}, we report the traditional VQA accuracy and observe that the Block variant of our(SG+QAA) model achieves higher accuracy than other state-of-the-art methods. Furthermore, the Concat-MLP model achieves almost same traditional VQA accuracy with or without QAA features ($\sim84.0$). Interestingly, one can notice that, even with same VQA accuracy, our model achieves a significant boost in both versions of Arithmetic and Harmonic MPT. These findings support our hypothesis that the QAA features encode salient object-level information that helps consider high-level visual concepts when answering difficult questions. 

\begin{table*}[ht!]
    \caption{Evaluation of our QAA models on the testset of TDIUC \cite{kafle2017analysis} dataset and comparison with state-of-the-art methods. The first 12 rows report the unnormalized accuracy for each question-type. The Arithmetic MPT and Harmonic MPT are unnormalized averages, and Arithmetic N-MPT and Harmonic N-MPT are normalized averages of accuracy scores for all question type. The last row shows the simple VQA accuracy for all models. Using complementary QAA features, the models ability to answer rare questions increased significantly (\ie higher Harmonic MPT and N-MPT) for all cases.}
    \centering
    \scalebox{.9}{%@{\extracolsep{2.5pt}}
    \begin{tabular}{lccc c>{\columncolor{clr4}}cc>{\columncolor{clr4}}cc>{\columncolor{clr4}}cc>{\columncolor{clr4}}c}
    \toprule
    \rowcolor{clr3} &MCB
    &NMN
    &RAU
    &\multirow{1}{*}{Linear}
    &Ours
    &\multirow{1}{*}{Concat}
    &Ours
    &\multirow{1}{*}{Mutan}
    &Ours
    &\multirow{1}{*}{Block}
    &Ours\\
    \rowcolor{clr3} &\cite{fukui2016multimodal}
    &\cite{andreas2016neural}
    &\cite{noh2016training}
    &&(SG+QAA)&-MLP&(SG+QAA)&&(SG+QAA)&&(SG+QAA)\\
    \cmidrule(lr){1-1} \cmidrule(lr){2-4} \cmidrule(lr){5-6} \cmidrule(lr){7-8} \cmidrule(lr){9-10} \cmidrule(lr){11-12}
    Scene Recog.            &93.0   &91.9   &94.0   &50.9   &93.1   &92.5   &93.0   &92.2   &92.4   &92.8   &92.8\\
    Sport Recog.            &92.8   &90.0   &93.5   &19.0   &93.7   &93.4   &94.1   &93.0   &92.9   &93.5   &93.5\\
    Color Attributes        &68.5   &54.9   &66.9   &55.7   &67.1   &65.4   &68.2   &66.3   &66.2   &68.6   &64.5\\
    Other Attributes        &56.7   &47.7   &56.5   &0.1    &54.9   &56.3   &56.4   &52.1   &52.4   &57.9   &56.1\\
    Activity Recog.         &52.4   &44.3   &51.6   &0.0    &50.9   &52.3   &53.0   &49.3   &50.2   &53.2   &52.4\\
    Pos. Reasoning          &35.4   &27.9   &35.3   &7.3    &33.4   &32.2   &35.4   &29.4   &29.9   &36.1   &34.7\\
    Sub-Obj Recog.          &85.4   &82.0   &86.1   &23.8   &85.7   &86.1   &86.5   &85.2   &85.5   &86.2   &85.9\\
    Absurd                  &84.8   &87.5   &96.0   &90.3   &88.2   &92.4   &92.4   &90.0   &89.1   &90.7   &92.1\\
    Object Utility          &35.0   &25.1   &31.6   &15.2   &29.3   &26.2   &35.7   &27.4   &30.4   &34.5   &37.4\\
    Object Presence         &93.6   &92.5   &94.4   &93.5   &94.3   &94.3   &94.4   &93.8   &93.9   &94.1   &94.2\\
    Counting                &51.0   &49.2   &48.4   &50.1   &51.2   &53.0   &52.6   &51.2   &50.4   &51.1   &51.2\\
    Sentiment Undstd.       &66.3   &58.0   &60.1   &56.3   &65.8   &65.7   &66.3   &63.2   &61.0   &66.0   &63.5\\
    \midrule
    Arithmetic MPT          & 67.9   &62.6   &67.8   &38.5   & 68.3(29.8$\uparrow$)   &67.6   & \textbf{69.0}(1.4$\uparrow$)   &66.2   &66.3(0.1$\uparrow$)   &68.4   &68.8(0.4$\uparrow$)\\
    Harmonic MPT            &60.5   &51.9   &59.0   &0.0    &58.1(58.1$\uparrow$)   &57.3   & \textbf{61.3}(4.0$\uparrow$)   &55.1  &56.7(1.6$\uparrow$)   &60.0  &61.1(1.1$\uparrow$)\\
    \cmidrule(lr){5-6} \cmidrule(lr){7-8} \cmidrule(lr){9-10} \cmidrule(lr){11-12}
    Arithmetic N-MPT        & 42.5   &34.0   &41.0   &29.8   & 54.1(34.3$\uparrow$)   &53.4   & \textbf{56.4}(3.0$\uparrow$)   &53.1  &53.7(0.6$\uparrow$)   &54.7  &55.9(1.2$\uparrow$)\\
    Harmonic N-MPT          &27.3   &16.7   &24.0   &0.0    &32.3(32.3$\uparrow$)   & 28.2   & \textbf{38.8}(3.0$\uparrow$)   &32.3  &32.8(0.6$\uparrow$)   &34.1  &38.2(1.2$\uparrow$)\\
    \cmidrule(lr){5-6} \cmidrule(lr){7-8} \cmidrule(lr){9-10} \cmidrule(lr){11-12}
    Simple Accuracy         &81.9   &79.6   &84.3   &72.9   &82.6(9.7$\uparrow$)   &83.9   &84.0(0.1$\uparrow$)   &82.5  &82.7(0.2$\uparrow$)   &84.5  &\textbf{84.6}(0.1$\uparrow$)\\
    \bottomrule
    \end{tabular}
    % \vspace{-1.0em}
    }
    \label{tab:TDIUC_results}
\end{table*}
%%%%%%%%%%%%%%%%%%%%%%%%%%%%%%%%%%%%%%%

\subsection{Inference with Global Representation}
\label{sec:inf_global_rep}
\begin{figure}[!t]
\centering
%\vspace{-1em}
      \includegraphics[width=.8\linewidth]{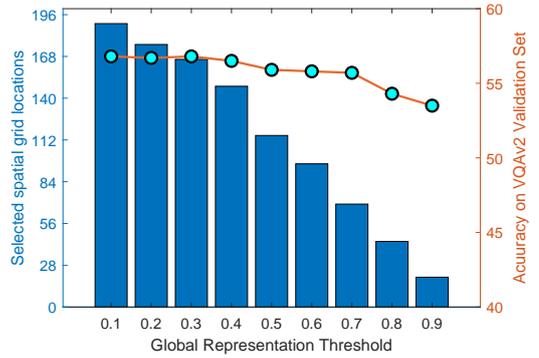}              \vspace{-.8em}
      \caption{{VQA accuracy (right y-axis) using complementary Image-Question-Agnostic Attention (IQAA) Features. IQAA feature is generated by selecting a global representation threshold (x-axis) and corresponding spatial grid locations (left y-axis).}
      }
      \label{fig:global_thresh}
\end{figure}

\begin{figure}[!t]
	\centering
	%\vspace{-1em}
          \includegraphics[width=.7\linewidth]{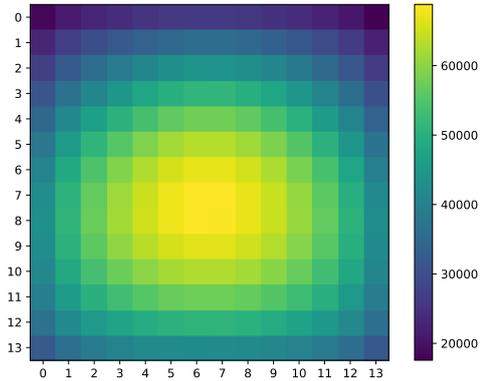}              %\vspace{-.8em}
          \caption{{Global Representation of \textit{object maps}: Count of object presence at each of the $196~(14 {\times} 14)$ spatial grid locations generated from the training images of VQA dataset.}} 
          \label{fig:global_rep}
\end{figure}

We further experiment with Image-Question-Agnostic Attention (IQAA) where the attention feature is generated without looking at the input question \textit{and} image. To do so, \textbf{first}, we create a global representation of \textit{object maps} by counting object presence at each spatial grid location for all images in the dataset. In Fig.~\ref{fig:global_rep}, we show such a global representation from the count of object presence, $\mathcal{C} \in \mathbb{R}^{14 \times 14}$, of VQA dataset training images (\ie COCO trainset 2014 images) on a $14 \times 14$ grid. We can see from this figure that most objects present in an image occupy the center grids. We leverage this centre bias to create fixed \textit{object maps}, that in turn is used to generate IQAA features. \textbf{Second}, the count vector is min-max normalized between $[0,1]$ (x-axis of Fig.~\ref{fig:global_thresh}). The left y-axis shows the number of grid locations selected when applying different thresholds on the normalized count measures. It ranges from $191$ to $22$ grid locations when the threshold is varied between $0.1$ to $0.9$. \textbf{Third}, we treat the selected grid location for a set threshold as fixed \textit{object maps} and apply fixed map on the input visual feature as discussed in Sec.~\ref{sec:ins_att} for generating IQAA features. These IQAA features can be used instead of QAA features in a similar fashion to train any VQA model. 

%%%%%%%%%%%%%%%%%%%%%%%%%%%%%%%%%%%%%%%
%%%%%%%%%%%%%%%%%%%%%%%%%%%%%%%%%%%%%%%
%%%%%%%%%%%%%%%%%%%%%%%%%%%%%%%%%%%%%%%
%%%%%%%%%%%%%%%%%%%%%%%%%%%%%%%%%%%%%%%

%%%%%%%%%%%%%%%%%%%%%%%%%%%%%%%%
\begin{figure*}[!t]
\begin{center}
% \fbox{\rule{0pt}{2in} \rule{.9\linewidth}{0pt}}
\includegraphics[width=.9\linewidth, clip=true, trim=0cm 2.15cm 0cm 0cm]{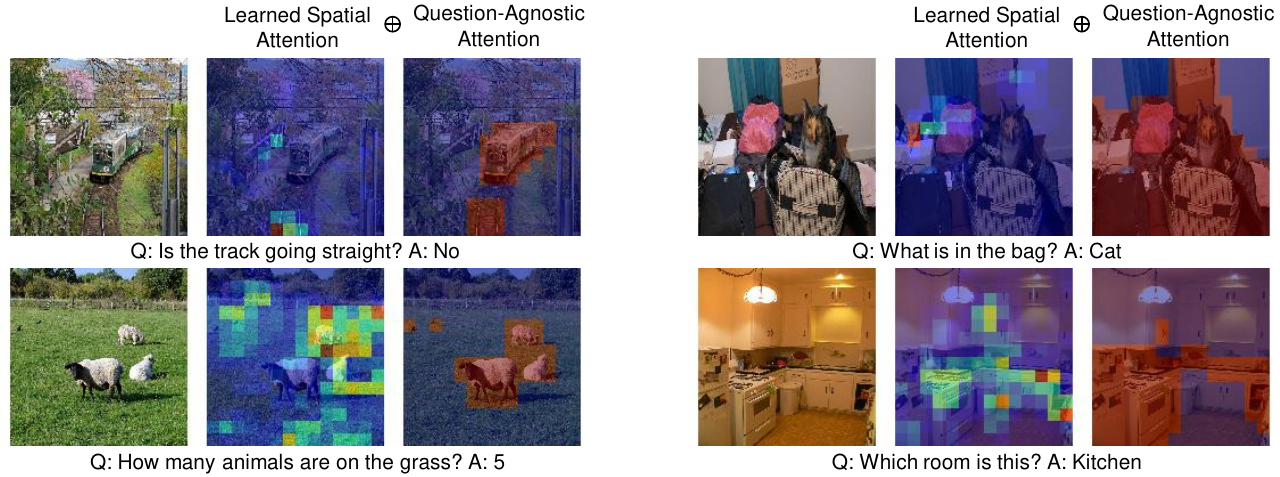}
\end{center}
\vspace{-2em}
  \caption{Qualitative results on VQAv2 val-set to demonstrate the effectiveness of using complementary QAA. The learned spatial attention ($2^{nd}$ and $5^{th}$ columns) focuses on regions  with or without objects, but QAA map is focused on objects.}
\label{fig:qual_results}
\end{figure*}
%%%%%%%%%%%%%%%%%%%%%%%%%%%%%%%%

\noindent \emph{By only using complementary IQAA features VQA models achieve reasonable performance.} We report the VQA accuracy score on VQAv2 validation set using our Block baseline model without spatial attention on the right y-axis of Fig.~\ref{fig:global_thresh}.
When using IQAA features, the VQA accuracy ranges from $53$ to $56$ when the global representation threshold is varied between $0.1$ to $0.9$. This means if one selects a fixed set of $24$ spatial grid locations at the center of the image, and trains a state-of-the-art VQA model with visual features of only these grid locations; the model can still achieve VQA accuracy comparable to when it looks at the whole image.
A similar finding was reported by Judd \textit{et al.} \cite{judd2009learning} where they show that humans tend to focus the object at the center when they take picture. Our finding further adds to that notion of \textit{Center Prior} by showing that humans also tend to ask questions about objects that are at the center of the image. By modeling the object presence prior in the dataset, one can achieve reasonable performance, without considering image specific object map. 
% Even though we run inference with pre-trained Mask-RCNN to generate QAA features as a light-weight prepossessing step, by modeling the object presence prior in the dataset, one can 
% further reduce pre-training computational burden by replacing QAA with IQAA, and
% achieve reasonable performance, without considering image specific object map. 

\subsection{Evaluation on the VQAv2 Testset}
\label{sec:eval_on_VQAv2_testset}
We evaluate our model's performance on the VQAv2 Test server and report accuracy for different question types on {the} Test-dev and Test-std dataset to compare with other contemporary state-of-the-art VQA models. For fair comparison, in Tab.~\ref{tab:VQAv2_testset} (1), we separate models that use spatial grid features (\ie visual features extracted by ResNet) and compare it with our SG+QAA model; in (2) the models that use Bottom-Up \cite{Anderson2017up-down} features and we compare our BU+QAA model. For both cases, our question-agnostic models employ Block fusion to jointly embed image and question features with a spatial attention mechanism. From Tab.~\ref{tab:VQAv2_testset}, we can see that when the QAA features are used alongside spatial grid features, the gain is more, compared to when used with BU features. As the BU features are a collection of top object bounding box features generated using Faster-RCNN, it also offers some object-level information to the VQA model. Thus, when used in combination with BU features, the overall performance gain in relatively small. However, as the question-agnostic features encompass the \textit{Object Map} of an image, it somewhat encodes the global spatial relationship between object and count information of object instances; it provides accuracy gain when answering \textit{Number} (\ie \textit{`How many?'}) question ($0.2\% \uparrow$ in test-standard). Overall, if a parallel branch trained using question-agnostic features is added to an existing VQA model, accuracy of the model increases.

\subsection{Qualitative Results}
\label{sec:qual_resutls}
We present qualitative results of our SG+QAA model with Block fusion in Fig.~\ref{fig:qual_results} to showcase the effectiveness of having complementary question-agnostic features. In the second row, first example, the model is tasked with a count question, asking \textit{`How many animals are on the grass?'} The learned spatial attention map is scattered in different image locations whereas the question-agnostic feature localizes five object instances that help the model answer correctly. Altogether, from the qualitative results, we can deduce that learned and question-agnostic attention provides complementary information which can be leveraged by VQA models to be able to correctly answer intelligent questions.

%-------------------------------------------------------------------------
\section{Conclusion}
\label{sec:conclusion}
In this paper, we introduced Question-Agnostic Attention that can be used to augment existing VQA approaches. Rather than using computationally intensive methods to learn question-specific attention, our approach derives attention only from the image, based on the insight that questions generally relate to object instances. We use an object parsing model to automatically generate an \textit{Object Map}, that has the same resolution as the feature map from a pre-existing classification network. The \textit{Object Map} is used to mask the convolutional feature map to generate question-agnostic attention features. When high-performing computationally-intensive VQA models are augmented with QAA, it improves their accuracy to be a new state-of-the-art. When simple linear models are augmented with QAA, they preform significantly better when answering question that require  {a} higher level of visual reasoning (\eg activity recognition), which a simplistic model cannot learn on its own. This capability provides the simplistic (low-complexity) models a significant boost that brings them close to state-of-the-art.

\bibliographystyle{IEEEtran}
\bibliography{egbib}

\end{document}